\documentclass[letterpaper, 10 pt, conference]{ieeeconf}
\IEEEoverridecommandlockouts
\usepackage{amsmath,amsfonts}
\usepackage{algorithmic}
\usepackage{array}
\usepackage[caption=false,font=normalsize,labelfont=sf,textfont=sf]{subfig}
\usepackage{textcomp}
\usepackage{stfloats}
\usepackage{hyperref}
\usepackage{url}
\usepackage{verbatim}
\usepackage{graphicx}
\usepackage{tikz}
\usepackage{cite}
\def\BibTeX{{\rm B\kern-.05em{\sc i\kern-.025em b}\kern-.08em
    T\kern-.1667em\lower.7ex\hbox{E}\kern-.125emX}}
\usepackage{balance}
\begin{document}
\title{Micro-Swarm Locomotion Optimization in Dynamic Flow using Multi-Objective Multi-Agent Reinforcement Learning}
\author{Josef Berman and Oren Gal 
\thanks{The Hatter Department of Marine Technologies, Leon H. Charney School of Marine Sciences, University of Haifa, Israel. Corresponding author: ds@josefberman.com}}

\markboth{}%
{}

\maketitle

\begin{abstract}
Coordinating micro-robotic swarms in realistic, time-dependent fluid environments remains a major challenge for biomedical and environmental applications. We present a hybrid CFD-MO-MARL (Computational Fluid Dynamics-Multi Objective-Multi Agent Reinforcement Learning) framework that couples a high-fidelity incompressible Navier--Stokes solver with decentralized proximal policy optimization to learn swarm control policies in oscillatory flow. Sixteen magnetically actuated micro-robots were simulated to navigate a pulsatile arterial waveform within a 2 mm channel while jointly optimizing upstream progression, energy efficiency, and motion smoothness. Conflicting objectives are resolved using Projected Conflicting Gradient (PCGrad) surgery. Without PCGrad, energy and smoothness rewards collapse during training, demonstrating that gradient conflict resolution is essential for stable multi-objective learning.

The converged policy achieves progress rewards of 6.5–7.0, energy efficiency of 0.63–0.65, and smoothness of 0.97–0.99, outperforming brute-force baselines by more than 8 reward units on the primary objective. Training reveals three emergent behaviors not encoded in the reward function: hydrodynamic throttling formations that reduce peak flow velocities, a cycle-synchronized ratchet mechanism that exploits flow reversals for upstream movement, and individualized final-approach strategies near the target boundary. These results demonstrate that physically realistic fluid--agent interactions can be integrated directly into multi-objective reinforcement learning, providing a scalable framework for micro-swarm control in biomedical navigation, environmental monitoring, and microfluidic systems.
\end{abstract}


\section{Introduction}

\PARstart{C}{oordinating} micro-robotic swarms in time-dependent fluid environments remains a major challenge for biomedical and environmental applications. At microscopic scales, viscous forces dominate inertial effects, requiring non-reciprocal actuation to generate propulsion \cite{Purcell1977}. In oscillatory flows such as pulsatile blood vessels or cyclically driven pipelines, micro-swimmers experience flow reversals, transient boundary layers, and evolving shear fields that can impede navigation and destabilize swarm formations \cite{Cermatori2014}. Hydrodynamic interactions further introduce time-varying wake effects that influence collective behavior \cite{Cheang2016}. Developing control strategies that simultaneously maintain directional progress, energy efficiency, and smooth motion under these conditions remains an open problem.

\subsection{Limitations of Existing Control Methods}

Traditional micro-swimmer control relies on global magnetic, acoustic, or optical actuation combined with model-based planning methods such as MPC \cite{Williams2017,diller2013independent}. While effective in steady flows, these approaches require accurate system models and continuous state feedback, assumptions that often fail in unsteady physiological environments \cite{martel2014magnetic}. Decentralized bio-inspired strategies improve scalability but can become trapped in oscillatory flow patterns and generally lack mechanisms for anticipating flow dynamics or coordinating large swarms \cite{dorigo2014swarm}.

Reinforcement learning (RL) offers an alternative by learning control policies directly through interaction with the environment \cite{sutton1998reinforcement}. Model-free methods such as PPO have demonstrated success in fluid locomotion and flow-control tasks without requiring explicit system identification \cite{novati2019controlled}. However, micro-swarm navigation introduces additional challenges associated with multi-agent coordination and competing objectives. Multi-agent reinforcement learning (MARL) addresses decentralized cooperation \cite{zhang2021multi}, while multi-objective reinforcement learning (MORL) enables simultaneous optimization of conflicting goals \cite{Hayes2022}. Gradient interference between objectives remains a major obstacle, motivating approaches such as Projected Conflicting Gradient (PCGrad), which mitigates destructive interactions during learning \cite{yu2020gradient}.

\subsection{Research Gap}

To the best of our knowledge, despite advances in RL, MARL, and CFD-based learning, no prior work has integrated high-fidelity computational fluid dynamics with multi-objective multi-agent reinforcement learning for swarm navigation in oscillatory flows. Existing CFD-RL studies primarily focus on single-agent systems, steady-flow environments, or single-objective optimization \cite{novati2019controlled,viquerat2021direct}. Consequently, the effects of time-dependent fluid dynamics on decentralized swarm coordination and multi-objective control remain largely unexplored.

\subsection{Contributions}

This work presents a hybrid CFD-MO-MARL framework that directly couples an incompressible Navier--Stokes solver with decentralized PPO policies for micro-swarm control in pulsatile flow environments. Sixteen magnetically actuated micro-robots jointly optimize upstream progression, energy efficiency, and motion smoothness. To address conflicting objectives, policy updates are reconciled using PCGrad, preventing destructive gradient interference during training. The proposed framework demonstrates that physically realistic fluid--agent interactions can be incorporated directly within reinforcement learning loops, enabling scalable and decentralized swarm control in dynamic fluid environments. These results establish a foundation for future applications in targeted biomedical navigation, environmental monitoring, and industrial microfluidics.

\section{Methods}

\subsection{Physical Setup}

\noindent All CFD simulations were performed in the open-source PhiFlow framework \cite{holl2024phiflow}, using its incompressible Navier--Stokes solver with semi-Lagrangian advection and projection-based pressure correction on a uniform Cartesian grid. Time integration used an explicit second-order scheme with a constant stable time step. No-slip conditions were imposed on solid boundaries, zero-gradient conditions in the transverse directions, and velocity and pressure fields were recorded at regular intervals.

\begin{figure}[!t]
    \centering
    \includegraphics[width=3.5in]{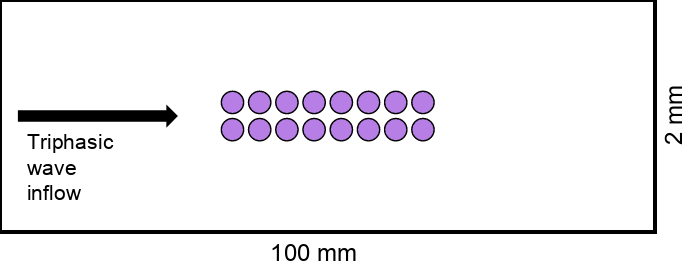}
    \caption{Initial physical setup. The domain is a $100~mm \times 2~mm$ tube containing a $2\times8$ grid of spherical swarm members with radius $0.25~mm$. A triphasic inflow waveform enters from the left boundary.}
    \label{fig:domain}
\end{figure}

\begin{figure}[!t]
    \centering
    \includegraphics[width=20pc]{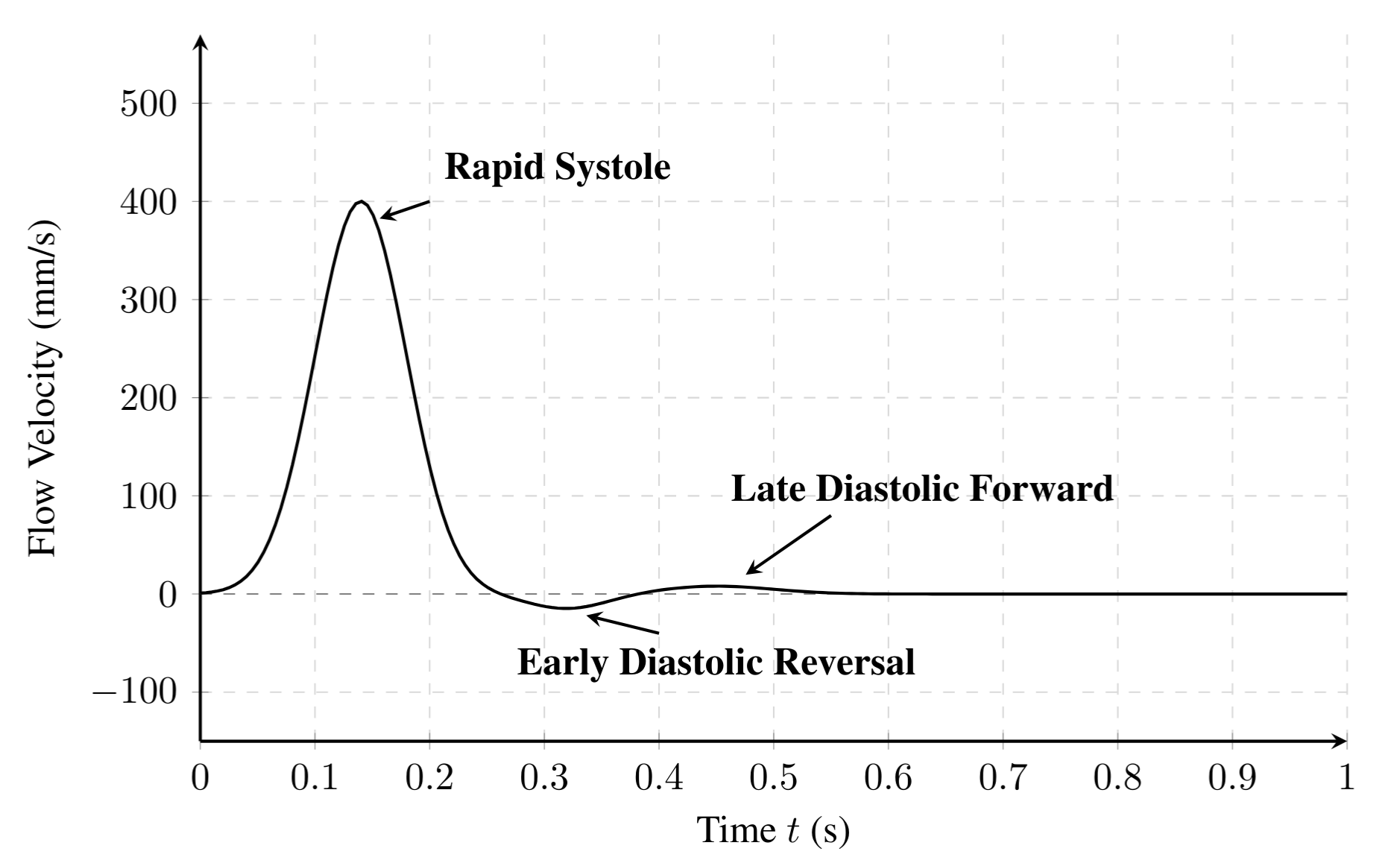}
    \caption{Triphasic temporal velocity profile, representing the rapid systole, early diastolic flow reversal, and late diastolic forward flow characteristic of a healthy peripheral artery with a 400 mm/s peak systolic velocity.}
    \label{fig:temporal-profile}
\end{figure}

The simulated domain represents a two-dimensional cross-section of a straight blood-filled tube with diameter $D=2\;mm$, density $\rho=1060\;kg\,m^{-3}$, and dynamic viscosity $\mu=3\times10^{-3}\;Pa\,s$. To approximate an effectively infinite tube, the computational length was set to $100\;mm$, with the outer $20\;mm$ at each end excluded from analysis to reduce boundary effects (Fig. \ref{fig:domain}). The flow was initialized as a developed parabolic profile and driven from the left inlet toward a right outlet held at atmospheric pressure. The inlet waveform followed a $1\;Hz$ triphasic arterial profile: a $400\;mm/s$ systolic peak during the first $150\;ms$, a $-15\;mm/s$ early-diastolic reversal for $100\;ms$, and an $8\;mm/s$ late-diastolic forward phase for the remainder of the cycle (Fig. \ref{fig:temporal-profile}). This corresponds to a laminar regime with $Re \approx 210$, while retaining unsteady inertial effects and transient local velocity gradients. Fluid motion is governed by the incompressible Navier--Stokes and continuity equations (Eqs. \ref{eq:NS}, \ref{eq:Continuity}).

\subsection{Micro-robot Dynamics and Force Coupling}
\noindent This research considers a swarm consisting of 16 swarm members (agents) made from a $L1_0$-FePt alloy arranged as a grid with spacing of $1\;mm$ in each direction in the middle of the tube. Each agent is modeled as a cross-section of a sphere of radius $r=250\;\mu m$, with mass $m=\frac{4}{3}\pi r^3\rho_{solid}$ ($L1_0$-FePt density $\rho_{solid}=15,120\;kg\,m^{-3}$). Although the fluid domain is two-dimensional, particle mass and magnetic actuation are computed for the corresponding three-dimensional spherical microrobots. Hydrodynamic forces arise from local pressure samples $p_k$ at four equispaced points around the cross-section of the sphere (the force was adjusted to the 2d-domain); the net force components in equation \ref{eq:forces}: 

\begin{equation}
\label{eq:forces}
    F_x=-\sum_{k=1}^4 \frac{\pi r}{2}p_k\cos{\theta_k},\;\;\;F_y=-\sum_{k=1}^4 \frac{\pi r}{2}p_k\sin{\theta_k}
\end{equation}

and each agent’s linear acceleration follows $m\ddot{\mathbf{x}}=\mathbf{F}_{hydro}+\mathbf{F}_{drag}+\mathbf{F}_{prop}$, where $\mathbf{F}_{hydro}=(F_x,\;F_y)$ is the hydrodynamic force due to pressure gradients, $\mathbf{F}_{drag}$ is the drag, and $\mathbf{F}_{prop}$ is the propulsive force generated magnetically and bounded by $\|\mathbf{F}_{prop}\|\leq850\;nN$ (assuming MRI-like gradient coils producing a gradient of $11.5\;mT/m$, and Saturation Magnetization of $L1_0$-FePt $M_{sat}=1,140\;kA/m$) \cite{Bai2024}.  Agents update their positions via explicit time integration of the acceleration.

\subsection{Governing Forces}

\noindent For this scenario, we consider four governing forces acting on or by each swarm member: hydrodynamic force (Eq. \ref{eq:NS}, \ref{eq:Continuity}) and drag force (Eq. \ref{eq:Drag}) acting on each member by the fluid, internal force (Eq. \ref{eq:Internal}) enacted by each member onto the fluid and other members, and contact forces (Eq. \ref{eq:Contact}) acting on each member by other members.

\begin{equation}
    \label{eq:NS}
    \rho\left(\frac{\partial \mathbf{u}}{\partial t}+(\mathbf{u}\cdot\nabla)\mathbf{u}\right)=-\nabla p+\mu\nabla^2\mathbf{u}+\mathbf{F}_{OSC}(A,\omega,t)
\end{equation}

\begin{equation}
    \label{eq:Continuity}
    \nabla\cdot\mathbf{u}=0
\end{equation}

where $\mathbf{u}$ is the velocity vector field, $p$ is the scalar pressure field, $\rho$ and $\mu$ are the density and the dynamic viscosity of the fluid, and $\mathbf{F}_{OSC}$ is an oscillatory driving force at the inlet imposing a trapezoidal waveform of amplitude $A$, frequency $\omega$, emulating physiological pulsatile or environmental oscillations.

\begin{equation}
    \label{eq:Drag}
    \mathbf{F}_{i,drag}=\frac{1}{2}\rho_{fluid}AC_d \|v_{i,rel}\|^2\hat{v}_{i,rel}
\end{equation}

where $\rho_{fluid}$ is the density of the fluid, $A$ is the contact area between the fluid and the $L1_0$-FePt sphere, $v_{i,rel}$ is the relative velocity between the fluid and swarm member $i$, calculated by sampling the fluid velocity around each swarm member, and $C_d$ is the drag coefficient, calculated using Eq. \ref{eq:C_d}:

\begin{equation}
    \label{eq:C_d}
        C_d=\begin{cases}
            \frac{24}{Re} & Re<0.1 \\
            \frac{24}{Re}\left( 1+0.15Re^{0.687} \right) & 0.1\le Re < 1000 \\
            0.44 & Re \ge 1000
        \end{cases}
\end{equation}

where $Re$ is the Reynolds number.

\begin{equation}
\label{eq:Internal}
    \mathbf{F}_{i,internal}=(a_{i,x},a_{i,y})
\end{equation}

where $a_{i,x}$ and $a_{i,y}$ are the force components of swarm member $i$, corresponding to the actions of the agent optimized using reinforcement learning techniques.

\begin{equation}
    \label{eq:Contact}
    \mathbf{F}_{i,contact}=\sum_{j\neq i}\left[(a_{j,x},a_{j,y})\cdot\hat{n}_{ij}\right]\hat{n}_{ij}
\end{equation}

where $a_{j,x}$ and $a_{j,y}$ are the force components of all swarm members $j$ \underline{in contact} with swarm member $i$.

\subsection{Numerical Discretization and Solver Pipeline}

\noindent The domain was discretized on a uniform Cartesian grid with $\Delta x=\Delta y=0.1\;mm$, selected to resolve near-wall gradients and vortex structures. Swarm forces were updated every $\Delta t=5\times10^{-3}\;s$, while the fluid solver used 20 substeps of $\Delta t_{\mathrm{sub}}=2.5\times10^{-4}\;s$ to satisfy the CFL stability constraint with Courant number approximately equal to 1. At each fluid substep, the velocity field was advanced by explicit viscous diffusion,
$\Delta\mathbf{u}=\nu\nabla^2\mathbf{u}\Delta t_{\mathrm{sub}}$, semi-Lagrangian advection, and projection-based incompressibility enforcement through the pressure Poisson equation $\Delta p=-\rho\nabla\cdot(\mathbf{u}\cdot\nabla\mathbf{u})$. This pipeline provides stable integration of the unsteady laminar flow while resolving the local fluid--swarm interactions relevant to control.

\subsection{Control Problem and Optimization Framework}

\noindent Micro-robotic swarm control is formulated as a multi-agent multi-objective Markov decision process (MA-MOMDP) and optimized using model-free PPO under a centralized-training, decentralized-execution (CTDE) paradigm.

\subsubsection{MDP Specification and Transition Dynamics}

\noindent At each timestep, the joint state $s\in S$ is formed by concatenating decentralized per-agent observations. Each agent observes an 8-dimensional local feature vector containing its position $(x,y)$, velocity $(v_x,v_y)$, and four pressure samples at equispaced points around its circumference. During training, the joint state is provided to the centralized critic to reduce multi-agent non-stationarity, while execution remains decentralized. Each agent $i$ selects a continuous two-dimensional propulsive force
$F_{i,\mathrm{internal}}=(a_{i,x},a_{i,y})$, bounded by the magnetic actuation limit $F_{\max}$. Transitions are governed by the coupled fluid--structure solver: the Navier--Stokes propagator updates the flow field, and agent motion is obtained by integrating hydrodynamic, drag, internal, and contact forces.

\subsubsection{Multi-Objective Reward Framework}

\noindent The reward is defined as a per-agent matrix of shape $(N,3)$, corresponding to progress, energy efficiency, and smoothness objectives, and is used with a three-head critic. For agent $i$, the progress reward encourages upstream displacement, assigns terminal success/failure rewards at $x<20$ and $x>80$, and includes a small idle penalty (Eq. \ref{eq:progressReward}). The energy reward measures alignment between applied force and displacement, thereby rewarding mechanically productive actuation (Eq. \ref{eq:energyReward}). The smoothness reward is the cosine similarity between consecutive actions, penalizing jittery control (Eq. \ref{eq:smoothnessReward}).

\begin{equation}
    \label{eq:progressReward}
    r_{\text{progress},i}=\begin{cases}
        +10 & x_{t,i}<20 \\
        100\frac{\tanh\left(3\mathbf{\Delta}\right)}{\tanh(3)}-0.01 & x_{t,i}\in[20,80],\;\Delta x_i<0 \\
        100\mathbf{\Delta}-0.01 & x_{t,i}\in[20,80],\;\Delta x_i\ge0 \\
        -10 & x_{t,i}>80
    \end{cases}
\end{equation}

\noindent where $\mathbf{\Delta}\equiv\Delta x_i/(x_\text{failure}-x_\text{success})$.

\begin{equation}
    \label{eq:energyReward}
    r_{\text{energy},i}=
    \frac{\mathbf{F}_i\cdot \Delta \mathbf{r}_i}
    {\mathbf{F}_{\text{max}}\|\Delta\mathbf{r}_i\|}
\end{equation}

\begin{equation}
    \label{eq:smoothnessReward}
    r_{\text{smooth},i}=
    \frac{\mathbf{a}_{i,t}\cdot \mathbf{a}_{i,t-1}}
    {\|\mathbf{a}_{i,t}\|\|\mathbf{a}_{i,t-1}\|}
\end{equation}

\subsubsection{Actor Network Architecture (Decentralized Execution)}

\begin{figure}[!t]
    \centering
    \includegraphics[width=2.5in]{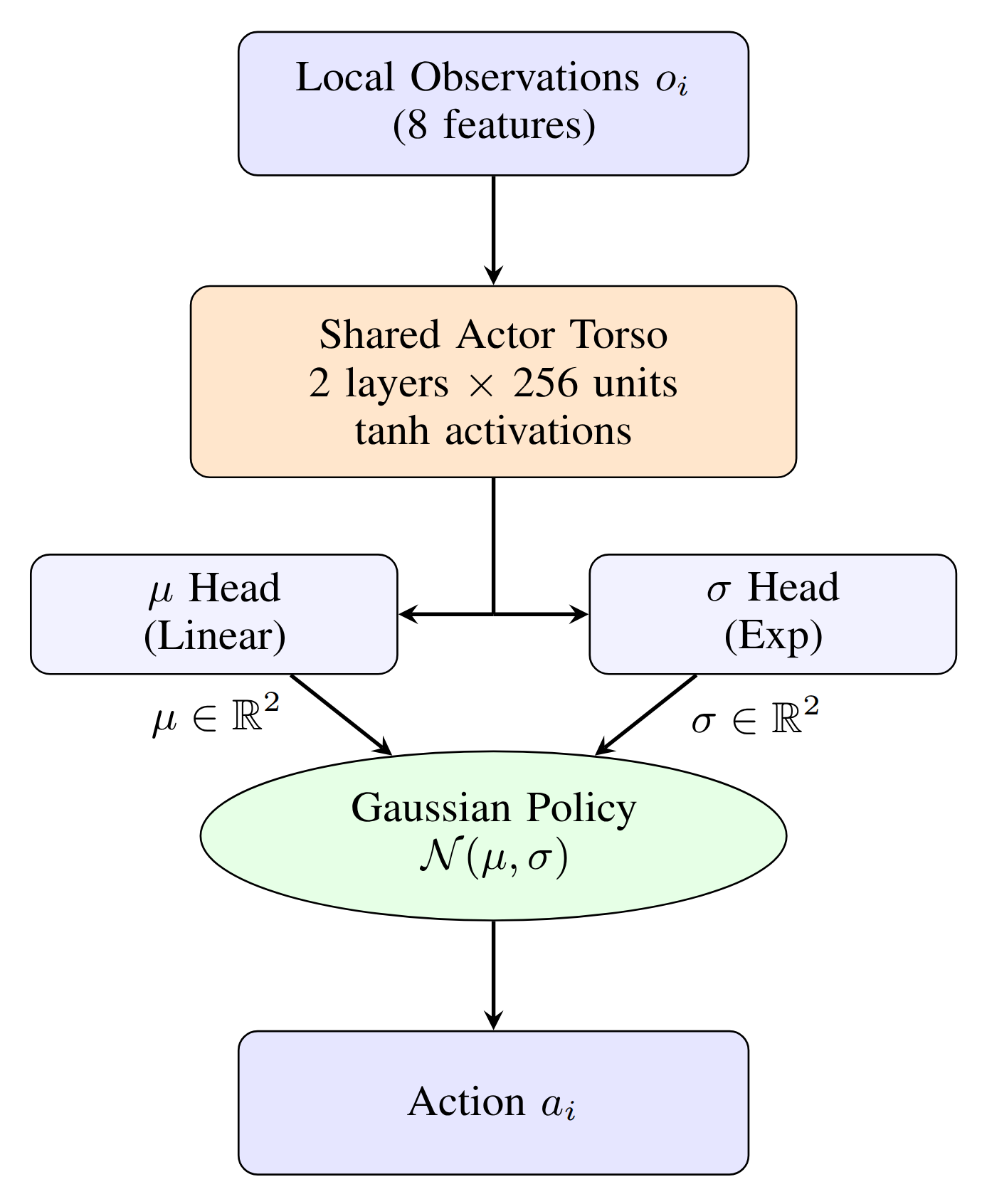}
    \caption{Architecture of the decentralized actor network. Each agent processes local observations through a shared torso to parameterize a Gaussian policy distribution.}
    \label{fig:actor_arch}
\end{figure}

The stochastic policy $\pi_{\theta}(a_i|o_i)$ is approximated using a shared actor torso that facilitates decentralized execution. Each agent $i$ processes only its local 8-dimensional observation $o_i$ through two hidden layers of 256 functional units each, utilizing $tanh$ activation functions. The network terminates in a Gaussian head that outputs the means ($\mu$) and log-standard deviations for the 2D action space. This factorized Gaussian approach allows the swarm to maintain adaptive heterogeneity while navigating unsteady flow fields (Fig. \ref{fig:actor_arch}).

\subsubsection{Critic Network Architecture (Centralized Training)}

\begin{figure}[!t]
    \centering
    \includegraphics[width=3.5in]{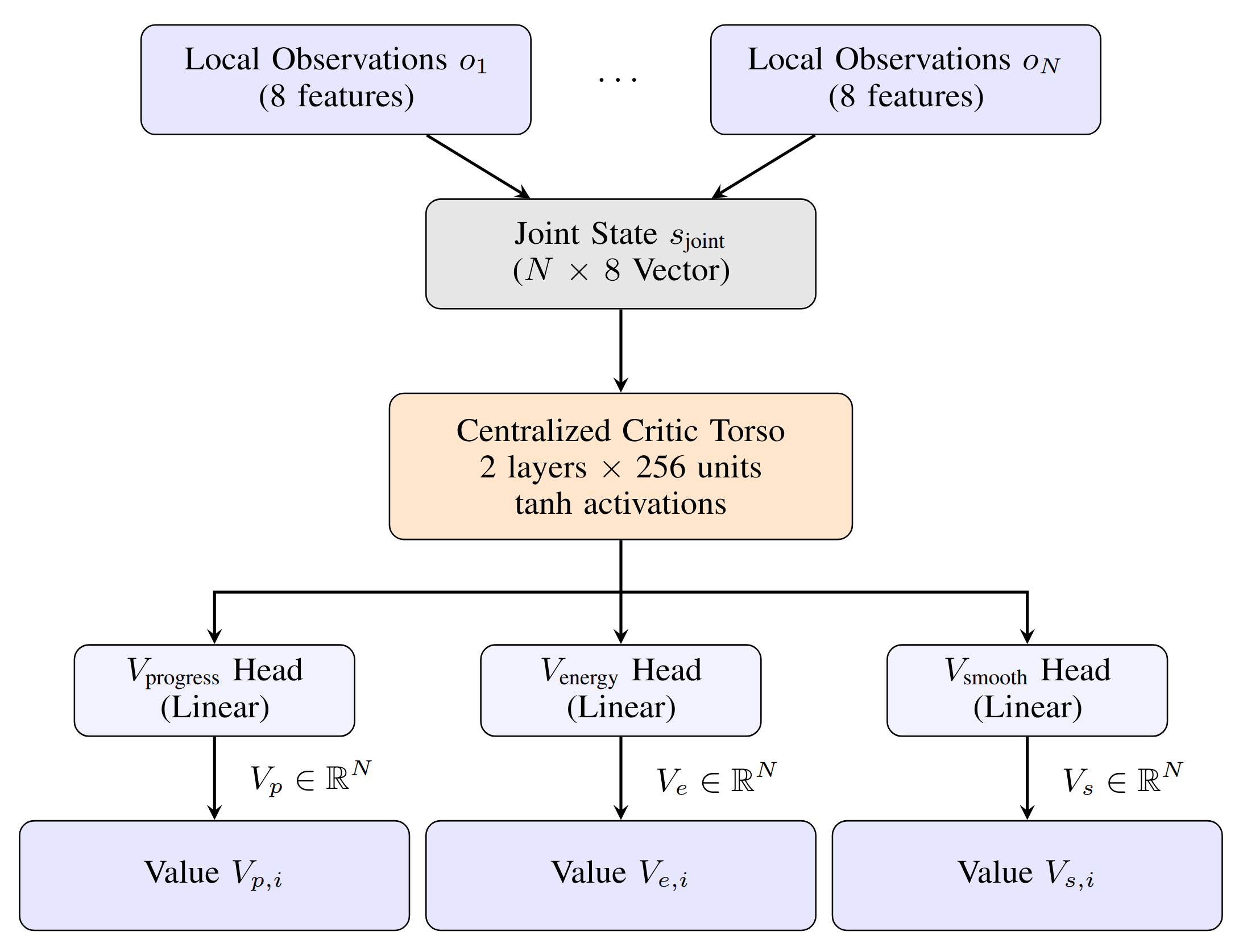}
    \caption{Architecture of the centralized multi-objective critic. The network processes the joint state of the swarm to produce independent value estimates for progression, energy efficiency, and smoothness objectives.}
    \label{fig:critic_arch}
\end{figure}

The centralized state-value function $V_{\phi}(s_{joint})$ is estimated by a critic network that leverages the joint observation of all agents ($N \times 8$). The critic utilizes the same fundamental MLP structure as the actor but employs three independent linear heads -- one for each objective (progress, energy, and smoothness). By processing the concatenated joint state, the critic provides a global baseline for the Generalized Advantage Estimation (GAE), which is essential for resolving credit assignment in the multi-agent context (Fig. \ref{fig:critic_arch}).

\subsubsection{Multi-Objective PPO Update with PCGrad}

\noindent We implement MOMAPPO as a modified PPO algorithm with decentralized stochastic actors, a centralized three-head critic, and objective-specific gradient surgery. During rollout collection, each environment returns a per-agent reward matrix
$\mathbf{R}_t\in\mathbb{R}^{N\times 3}$, whose columns correspond to progress, energy efficiency, and smoothness. The scalar Gym reward is used for environment compatibility and logging, while learning preserves separate objective channels through the reward buffer, GAE computation, critic heads, and actor-gradient computation.

For each rollout, the centralized critic predicts three per-agent value vectors,
$V_\phi^{p}(s_t)$, $V_\phi^{e}(s_t)$, and $V_\phi^{s}(s_t)\in\mathbb{R}^{N}$,
from the joint swarm state. Generalized Advantage Estimation is computed independently for each objective $k\in\{p,e,s\}$ and swarm member $i$:
\begin{equation}
    \hat{A}^{k}_{t,i} =
    \sum_{l=0}^{T-t-1}(\gamma\lambda)^l
    \left[
    r^{k}_{t+l,i}
    +\gamma V^{k}_{\phi,i}(s_{t+l+1})
    -V^{k}_{\phi,i}(s_{t+l})
    \right]
\end{equation}

where $r^{k}_{t+l,i}$ is the long-trajectory reward, $\gamma$ is the discount factor, $\lambda$ is the trajectory length factor.

The shared actor parameterizes a factorized Gaussian policy over all agents. The PPO likelihood ratio is computed from the summed joint log-probability over agents and action dimensions:
\begin{equation}
    \rho_t(\theta)=
    \exp\left[
    \log\pi_\theta(\mathbf{a}_t|\mathbf{o}_t)
    -
    \log\pi_{\theta_{\mathrm{old}}}(\mathbf{a}_t|\mathbf{o}_t)
    \right]
\end{equation}
For each objective, a separate clipped PPO actor loss is computed:
\begin{equation}
    L_\pi^{k}(\theta)=
    -\mathbb{E}_{t,i}
    \left[
    \min\left(
    \rho_t(\theta)\hat{A}^{k}_{t,i},
    \mathrm{clip}(\rho_t(\theta),1-\epsilon,1+\epsilon)\hat{A}^{k}_{t,i}
    \right)
    \right]
\end{equation}

PCGrad is applied to the three objective-specific actor losses,
$\{L_\pi^{p},L_\pi^{e},L_\pi^{s}\}$. Gradients are computed separately for each objective; if two gradients are negatively aligned, the conflicting component is removed by projection:
\begin{equation}
    \tilde{g}_i =
    \begin{cases}
    g_i - \frac{g_i^\top g_j}{\|g_j\|^2}g_j & g_i^\top g_j < 0\\
    g_i & \mathrm{otherwise}
    \end{cases}
\end{equation}
The projected gradients are averaged and assigned before the optimizer step. In the ablation study, this merge step of projected gradients is disabled and the three actor losses are combined by standard gradient summation.

The critic is trained using the summed value loss across the three objective heads:
\begin{equation}
    L_V(\phi)=
    \frac{1}{2}
    \sum_{k\in\{p,e,s\}}
    \mathbb{E}_{t,i}
    \left[
    \left(
    V_\phi^{k}(s_t)_i-\hat{R}^{k}_{t,i}
    \right)^2
    \right]
\end{equation}
and entropy regularization is added to maintain exploration. Thus, PCGrad resolves conflicts among objective-specific actor gradients, while the centralized critic provides independent per-objective, per-agent value baselines.

\noindent Training used 4 parallel Gym environments for 27,000 steps each. Rollouts contained 16 simulation steps per update, followed by 4 PPO epochs with mini-batches of 4 samples. Episodes terminated when the swarm mean position reached $x_{\mathrm{success}}=20~mm$, when any member crossed $x_{\mathrm{failure}}=80~mm$, or were truncated after 10~s or CFD divergence. Hyperparameters were $\epsilon=0.2$, entropy coefficient $0.01$, $\gamma=0.95$, $\lambda=0.95$, learning rate $10^{-3}$, and Adam optimization.

\subsection{Reproducibility and Data Management}

\noindent Physical parameters, swarm configurations, and RL hyperparameters are version-controlled. Flow fields, trajectories, and rewards are logged during training and post-processed into metrics and visualizations for reproducible comparison \cite{Berman2025}.

\section{Results}

\subsection{Multi-Objective Reinforcement Learning Convergence and Policy Stabilization}

\begin{figure}[!t]
    \centering
    \includegraphics[width=3.5in]{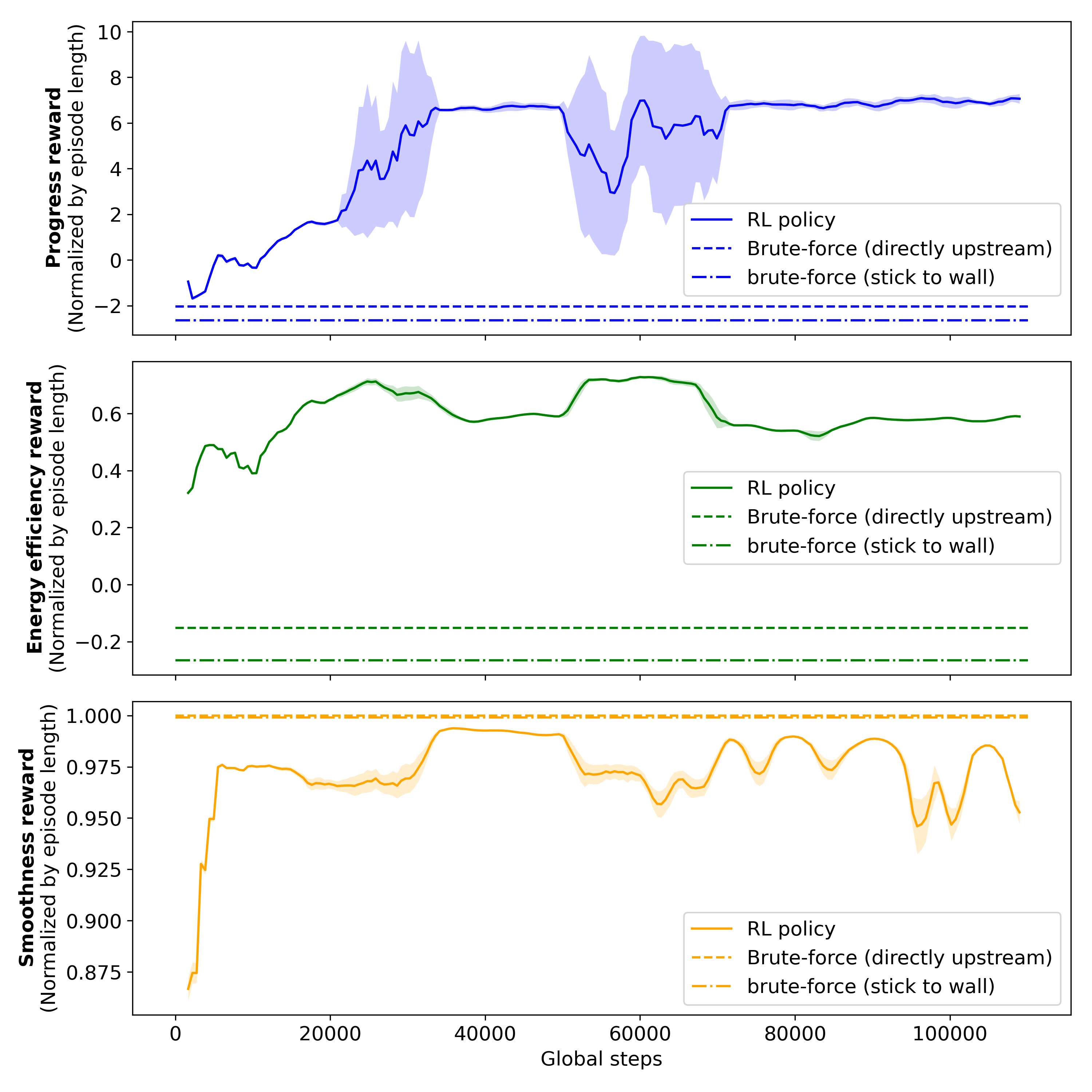}
    \caption{Training curves for the MOMAPPO policy versus direct-upstream and wall-hugging brute-force baselines. Curves show normalized per-episode rewards over global environment steps, with shaded regions denoting $\pm1$ standard deviation across parallel environments. From top to bottom: progress, energy efficiency, and action smoothness. MOMAPPO outperforms both baselines in progress and energy efficiency while maintaining near-maximal smoothness.}
    \label{fig:reward-per-episode}
\end{figure}

\noindent Fig. \ref{fig:reward-per-episode} shows the per-episode learning curves for progress, energy efficiency, and smoothness over approximately 110,000 global environment steps. Performance is compared against two brute-force baselines: constant upstream thrust and upstream motion biased toward the channel walls.

The progress reward exhibits the strongest learning signal. Starting near zero, the learned policy rapidly improves between 5,000 and 25,000 steps and converges to approximately 6.5–7.0, exceeding both baselines (approximately -2.0 and -2.5) by more than 8 reward units. Elevated variance between 30,000 and 70,000 steps suggests exploration of higher-reward behaviors before convergence and stabilization after roughly 75,000 steps.

Energy efficiency improves more smoothly, increasing from approximately 0.30 to a stable plateau of 0.63–0.65 by 20,000 steps. In contrast, both brute-force baselines remain negative (-0.15 and -0.27), indicating inefficient force application. The learned policy therefore achieves sustained upstream motion while maintaining positive energetic returns.

Smoothness converges fastest, rising from approximately 0.87 to 0.97–0.99 within the first 5,000 steps and remaining highly stable thereafter. Although the brute-force baselines attain the theoretical maximum value of 1.0 due to their fixed actions, the learned policy remains within 0.03 of this optimum while substantially outperforming both baselines on progress and energy efficiency. This indicates that the agent successfully balances adaptive control with action consistency.

Overall, the convergence profiles demonstrate that the MOMAPPO framework learns a stable Pareto-improving policy that simultaneously outperforms both brute-force strategies in progress and energy efficiency while maintaining near-optimal smoothness across parallel environment instances.

\subsection{Emergent Behavior}

\begin{figure*}[!t]
    \centering
    \includegraphics[width=6in]{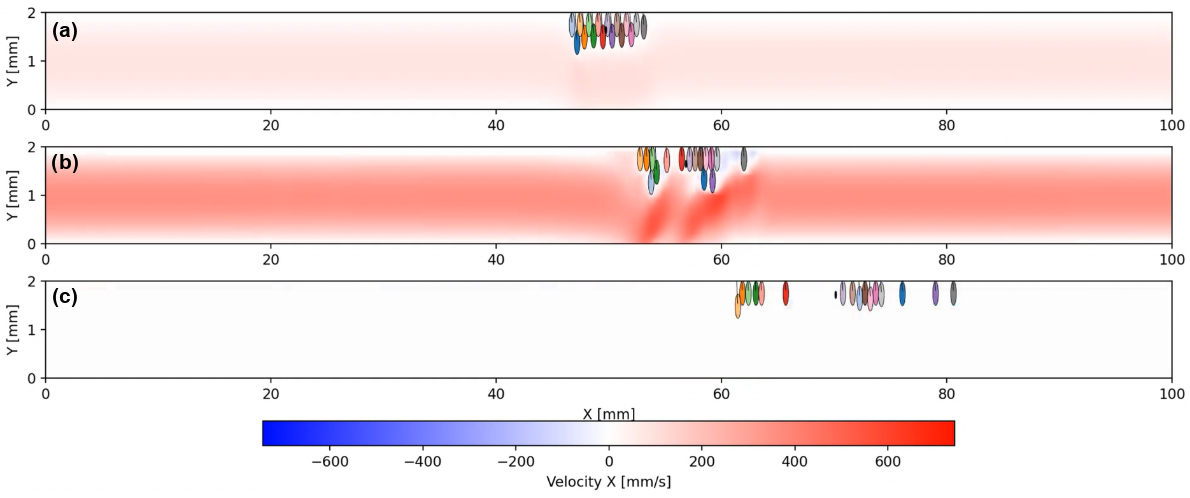}
    \caption{Snapshots show the axial velocity field $u_x$, swarm member positions, center of mass (black dot), and instantaneous action vectors for an untrained policy. Panels (a)–(c) show the uncontrolled failure mode: the initially compact swarm is displaced toward the upper wall, forms a narrow local constriction during peak forward flow, and generates high-velocity jets that accelerate downstream transport. Without learned coordination, the swarm does not exploit flow reversal or recover upstream displacement, and is instead advected toward the failure boundary.}
    \label{fig:before-learning}
\end{figure*}

\begin{figure*}[!t]
    \centering
    \includegraphics[width=6in]{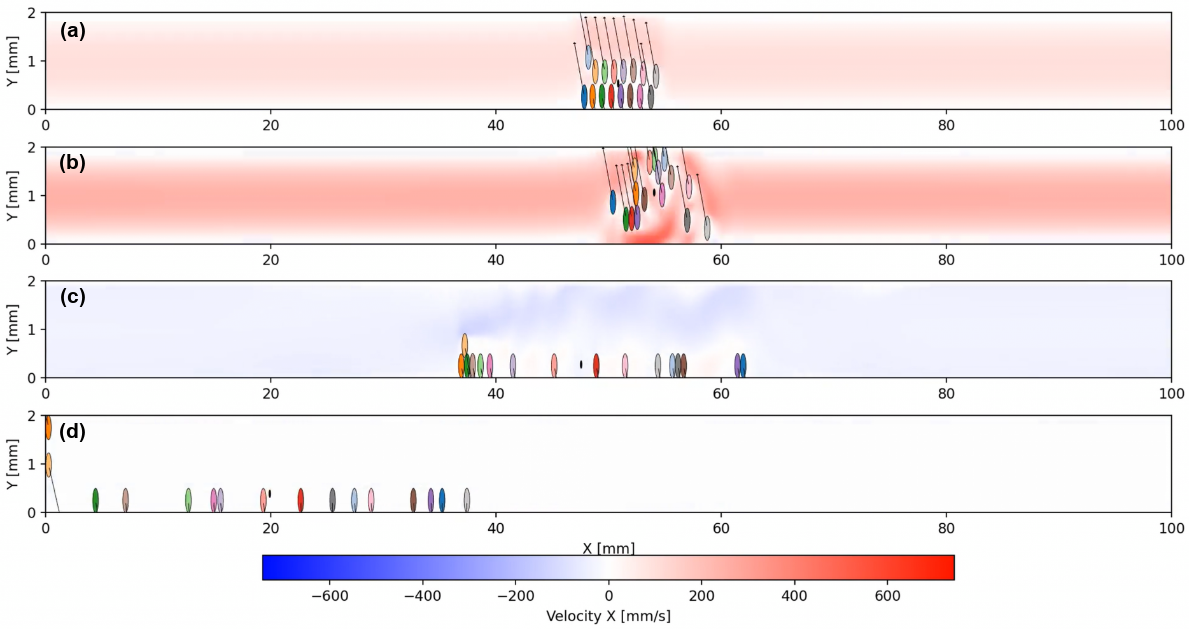}
    \caption{Snapshots show the axial velocity field $u_x$, swarm member positions, center of mass (black dot), and instantaneous action vectors for the trained MOMAPPO policy. Panels (a)–(b) show the learned collective response during forward flow, where the swarm remains organized and forms a distributed obstruction rather than collapsing into an uncontrolled constriction. Panel (c) shows the subsequent dispersed configuration as agents reconfigure across the channel, and panel (d) shows the late-stage individualized upstream approach as agents advance toward the success boundary. The sequence illustrates the learned transition from collective hydrodynamic interaction to ratchet-like recovery and individualized target acquisition.}
    \label{fig:after-learning}
\end{figure*}

\noindent The trained policy produces several collective behaviors that were not explicitly encoded in the reward function but emerged from the interaction between the multi-objective reward structure, pulsatile fluid dynamics, and decentralized control. Comparison of the untrained swarm (Fig. \ref{fig:before-learning}) and the trained policy (Fig. \ref{fig:after-learning}) reveals three distinct behavioral phases that exploit different portions of the flow cycle. To quantify the visually observed behavioral phases, we computed the swarm center of mass, axial and transverse dispersion, and radius of gyration over representative episodes. The center of mass was defined as $x_{COM}(t)=N^{-1}\sum_i x_i(t)$, while axial and transverse dispersions were computed as $\sigma_x(t)=\sqrt{N^{-1}\sum_i(x_i(t)-x_{COM}(t))^2}$ and $\sigma_y(t)=\sqrt{N^{-1}\sum_i(y_i(t)-y_{COM}(t))^2}$. The radius of gyration $R_g(t)=\sqrt{N^{-1}\sum_i\|\mathbf{r}_i(t)-\mathbf{r}_{COM}(t)\|^2}$, where $\mathbf{r}_i(t)=[x_i(t),y_i(t)]^\top$, was used as a scalar measure of overall swarm spreading. These metrics convert the visually observed phase sequence into measurable changes in swarm transport and organization.

\begin{figure*}[!t]
    \centering
    \includegraphics[width=6in]{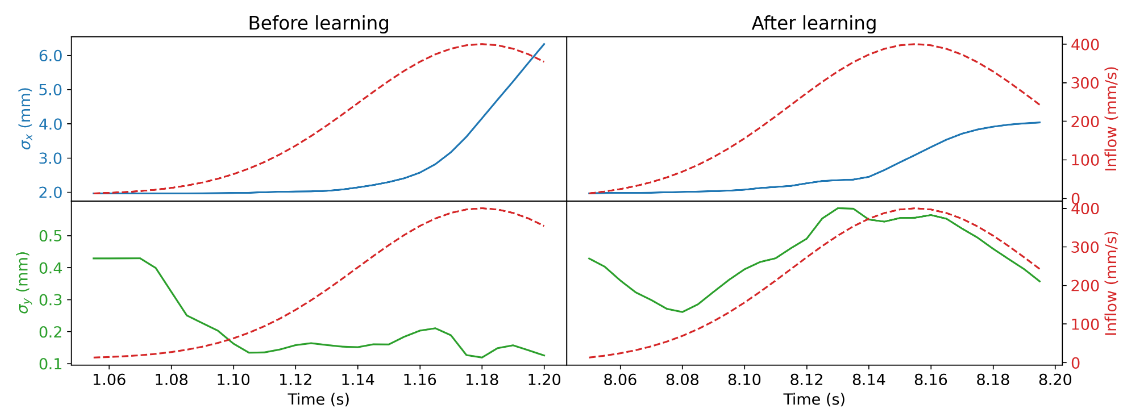}
    \caption{Swarm dispersion during representative forward-flow intervals before and after learning. Temporal evolution of axial dispersion $\sigma_x$ and transverse dispersion $\sigma_y$ is shown before and after learning, with the inlet waveform overlaid as a reference. Before learning, axial dispersion increases sharply during peak forward flow while transverse dispersion decreases, indicating collapse into a streamwise plume near the wall. After learning, axial dispersion grows more gradually, while transverse dispersion increases during the same interval, indicating that the trained swarm occupies more of the channel height and forms a cross-sectional obstruction. These dispersion signatures quantify the transition from passive downstream advection to learned hydrodynamic flow modulation.}
    \label{fig:dispersion}
\end{figure*}

\begin{figure*}[!t]
    \centering
    \includegraphics[width=6in]{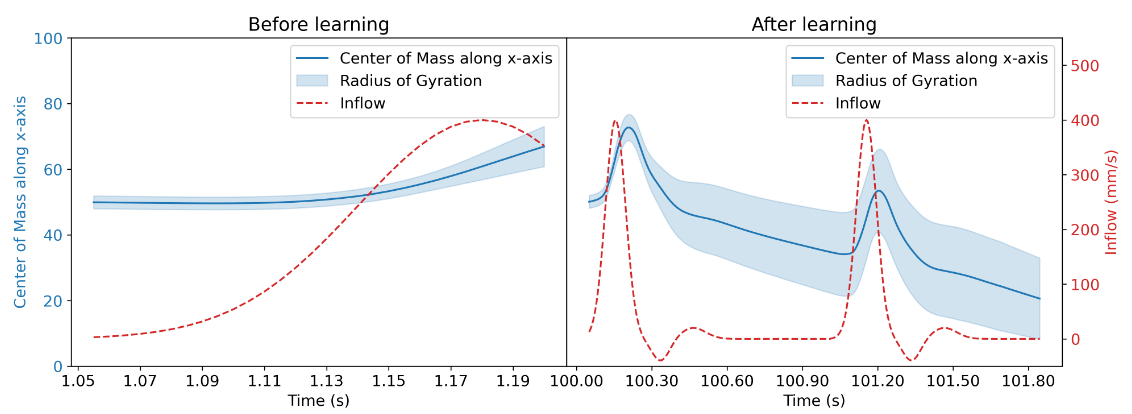}
    \caption{Center-of-mass transport and swarm-scale spreading before and after learning. The center-of-mass position $x_{COM}$, radius of gyration $R_g$, and inlet waveform are shown for representative untrained and trained episodes. Before learning, $x_{COM}$ increases monotonically during forward flow, indicating passive downstream transport toward the failure boundary, while $R_g$ increases as the swarm spreads axially. After learning, the center-of-mass trajectory becomes cycle-dependent: forward-flow pulses cause transient downstream displacement, but subsequent low- and reverse-flow intervals allow upstream recovery. The bounded $R_g$ envelope indicates that the trained policy maintains controlled swarm spreading while executing a ratchet-like transport strategy.}
    \label{fig:center-of-mass}
\end{figure*}

\subsubsection{Phase 1: Collective Flow Throttling During Peak Forward Flow}

During peak forward flow, the untrained swarm is displaced toward the upper wall and forms a narrow constriction, producing high local velocities and rapid downstream transport (Fig. \ref{fig:before-learning}). This failure mode is also reflected quantitatively by a sharp increase in axial dispersion $\sigma_x$ and a decrease in transverse dispersion $\sigma_y$, indicating collapse into a streamwise plume rather than cross-channel organization (Fig. \ref{fig:dispersion}). In contrast, the trained policy maintains a distributed near-wall configuration during forward flow (Fig. \ref{fig:after-learning}). The slower growth of $\sigma_x$, together with the increase in $\sigma_y$, indicates that the trained swarm occupies more of the channel height and acts as a hydrodynamic baffle. Rather than directly opposing the flow, the learned policy modulates the local flow passage while limiting uncontrolled downstream spreading.

\subsubsection{Phase 2: Coordinated Upstream Repositioning During Reverse Flow}

The trained trajectory exhibits cycle-dependent center-of-mass motion rather than monotonic downstream advection. In the untrained case, $x_{COM}$ increases during forward flow, consistent with passive transport toward the failure boundary (Fig. \ref{fig:center-of-mass}). After learning, forward-flow pulses still produce transient downstream displacement, but subsequent low- and reverse-flow intervals allow upstream recovery. This produces a cycle-synchronized ratchet mechanism in which downstream losses are limited during forward flow and upstream gains are accumulated during favorable waveform intervals. The concurrent evolution of $R_g$, $\sigma_x$ and $\sigma_y$ indicates that this recovery is achieved through swarm reconfiguration rather than rigid translation of the initial formation.

\subsubsection{Phase 3: Individualized Final Approach}

As the swarm approaches the success region, the collective organization observed during the throttling and ratcheting phases becomes less pronounced. The increase in swarm-scale spreading, reflected by $R_g$ and $\sigma_x$, indicates that the agents no longer maintain a compact collective formation throughout the episode (Fig. \ref{fig:center-of-mass}). Instead, individual members exploit local flow conditions and their own control actions to continue upstream progression. This transition is consistent with a final-approach regime in which the progress objective dominates over formation maintenance, and hydrodynamic cooperation provides diminishing benefit near the target boundary.

\subsection{Influence of Gradient Surgery on Learned Policy}

\begin{figure}
    \centering
    \includegraphics[width=3.5in]{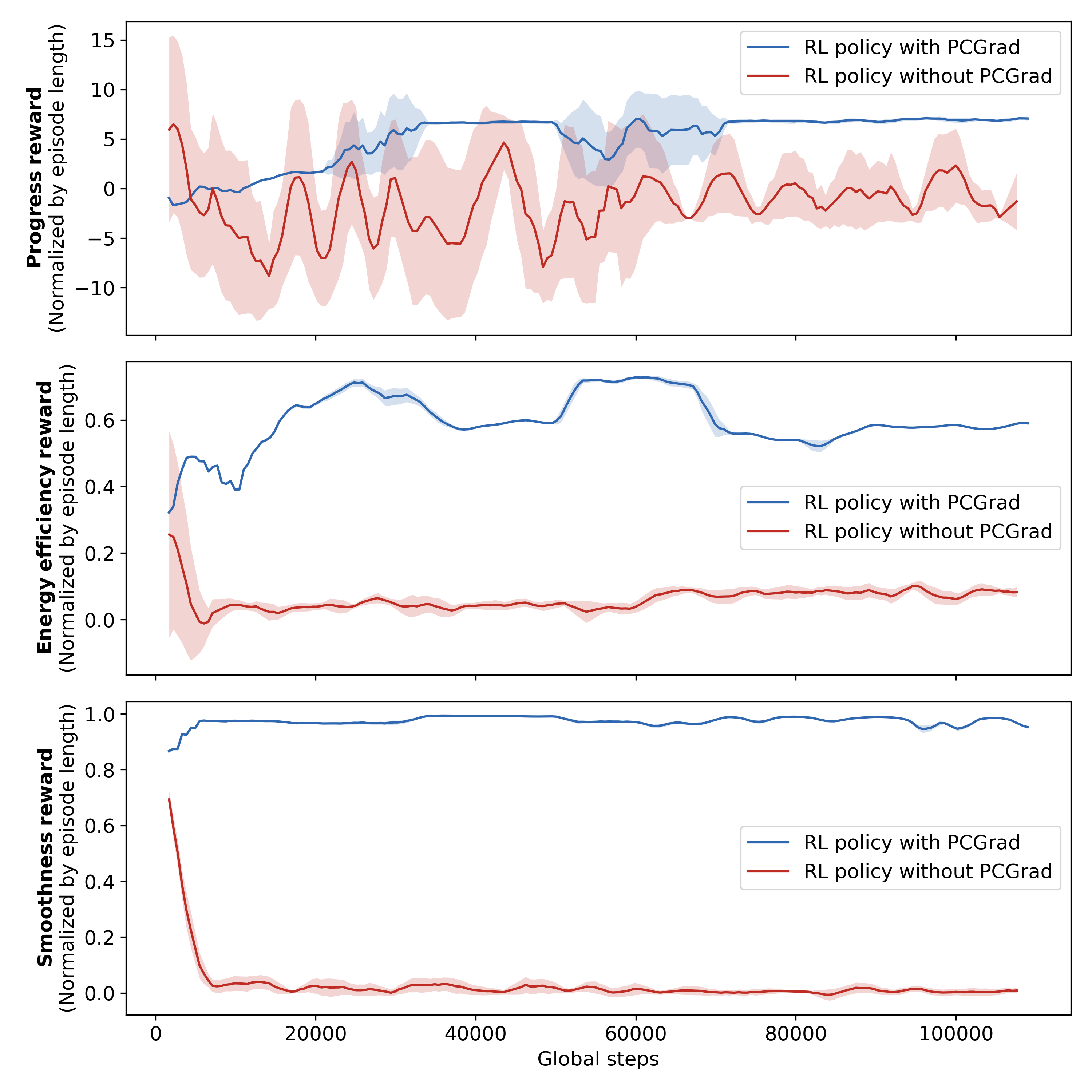}
    \caption{Ablation study comparing MOMAPPO with PCGrad (blue) and without PCGrad (red). Curves show normalized per-episode rewards over global environment steps; shaded regions denote $\pm1$ standard deviation across parallel environments. From top to bottom: progress, energy efficiency, and action smoothness. Without PCGrad, energy efficiency and smoothness collapse within 10,000 steps, while progress remains highly oscillatory, indicating destructive inter-objective interference.}
    \label{fig:with-without-pcgrad}
\end{figure}

\noindent To assess the contribution of PCGrad, we compare the full MOMAPPO policy with an ablated variant in which objective-specific gradients are replaced by standard summed-gradient updates, while keeping all other architectural and training settings fixed (Fig. \ref{fig:with-without-pcgrad}).

Removing PCGrad substantially degrades multi-objective stability. Without gradient surgery, the energy reward briefly improves during early training but collapses to near zero by approximately 10,000 steps, while the PCGrad policy maintains a stable plateau around 0.60--0.65. Smoothness exhibits an even stronger collapse: the ablated policy rapidly deteriorates to approximately zero, whereas the PCGrad policy converges to near-maximum smoothness ($\sim$0.97--1.00). These results indicate that naive gradient summation allows the progress objective to dominate the update direction, suppressing energy-efficient and temporally consistent control.

Progress is also less stable without PCGrad. The ablated policy exhibits persistent high-amplitude oscillations, with rewards ranging roughly from $-15$ to $+10$, whereas the PCGrad policy converges smoothly to approximately 6.5--7.0 with decreasing variance. Thus, although the ablated policy occasionally discovers useful behaviors, it fails to retain them under conflicting objective gradients. Overall, PCGrad preserves all three objective signals and is essential for stable multi-objective convergence in this setting.

\section{Discussion}

\noindent This work shows that coupling a high-fidelity Navier--Stokes solver with multi-objective multi-agent reinforcement learning enables stable and physically meaningful micro-swarm control in time-dependent flow. The learned policy does not rely on maximal upstream thrust, as in the brute-force baselines, but discovers a multi-phase ratchet strategy: passive anchoring during forward flow and active upstream repositioning during reverse flow. This behavior emerges despite the reward encoding only progress, energy efficiency, and smoothness, indicating that the coupled CFD--MARL environment enables non-trivial policy discovery rather than merely optimizing a prescribed maneuver.

A key emergent behavior is the two-layer throttling formation, which reduces peak local flow velocity from approximately $700\;mm/s$ to $400\;mm/s$. This suggests that the swarm learns to exploit collective hydrodynamic interaction and passive flow modulation instead of directly resisting the flow. The later transition from coordinated motion to individualized completion further indicates adaptive mode switching as agents approach the success boundary.

The ablation study demonstrates that PCGrad is essential for stable learning. Without gradient surgery, energy efficiency and smoothness collapse to near zero within 10,000 steps, while progress remains highly oscillatory. This indicates that naive gradient summation allows the dominant progress objective to suppress weaker secondary objectives, reducing the problem to unstable single-objective optimization. PCGrad therefore acts as a core algorithmic requirement rather than an optional enhancement in this CFD-MO-MARL setting.

The main limitations are the two-dimensional flow formulation, the assumption of a uniform magnetic gradient, and the fixed swarm size. Future work should extend the framework to 3D GPU-accelerated solvers \cite{kochkov2021machine}, spatially resolved actuation, graph-based policies for variable-size swarms \cite{jiang2018graph}, and validation under stenotic, irregularly pulsatile, or high-viscosity flows. More broadly, the results support direct integration of PDE-governed environments into multi-objective RL for biomedical microrobotics, environmental monitoring, and industrial microfluidics \cite{holl2020learning,kochkov2021machine}.

\section{Code Availability}
\noindent The complete code for this study can be found at \href{https://github.com/josefberman/FluxSwarm}{https://github.com/josefberman/FluxSwarm}.


\bibliographystyle{IEEEtran}

\bibliography{references}


\end{document}